\documentclass[10pt,twocolumn,letterpaper]{article}
\usepackage{tcolorbox} 

\usepackage[pagenumbers]{wacv} 

\definecolor{wacvblue}{rgb}{0.21,0.49,0.74}
\usepackage[pagebackref,breaklinks,colorlinks,allcolors=wacvblue]{hyperref}
\usepackage{comment}

\def\wacvPaperID{4} 
\def\confName{WACV}
\def\confYear{2026}

\title{PrismVAU: Prompt-Refined Inference System for Multimodal Video Anomaly Understanding}

\author{Iñaki Erregue$^{1,2,4}$\\
{\tt\small ierregal31@alumnes.ub.edu}
\and
Kamal Nasrollahi$^{3,4}$\\
{\tt\small kn@create.aau.dk}
\and
Sergio Escalera$^{1,2,3}$\\
{\tt\small sescalera@ub.edu}
\and
{\normalsize$^1$Universitat de Barcelona}
\and
{\normalsize$^2$Computer Vision Center}
\and
{\normalsize$^3$Aalborg University}
\and
{\normalsize$^4$Milestone Systems}
}

\begin{document}
\maketitle
\begin{abstract}
Video Anomaly Understanding (VAU) extends traditional Video Anomaly Detection (VAD) by not only localizing anomalies but also describing and reasoning about their context. Existing VAU approaches often rely on fine-tuned multimodal large language models (MLLMs) or external modules such as video captioners, which introduce costly annotations, complex training pipelines, and high inference overhead. In this work, we introduce PrismVAU, a lightweight yet effective system for real-time VAU that leverages a single off-the-shelf MLLM for anomaly scoring, explanation, and prompt optimization. PrismVAU operates in two complementary stages: (1) a coarse anomaly scoring module that computes frame-level anomaly scores via similarity to textual anchors, and (2) an MLLM-based refinement module that contextualizes anomalies through system and user prompts. Both textual anchors and prompts are optimized with a weakly supervised Automatic Prompt Engineering (APE) framework. Extensive experiments on standard VAD benchmarks demonstrate that PrismVAU delivers competitive detection performance and interpretable anomaly explanations---without relying on instruction tuning, frame-level annotations, and external modules or dense processing---making it an efficient and practical solution for real-world applications.
\end{abstract}
    
\section{Introduction}
\label{sec:intro}
Video Anomaly Detection (VAD) focuses on automatically identifying unusual or abnormal events, behaviors, or patterns in video data that deviate from what is considered ``normal'' or expected. It is a crucial task in computer vision with a wide range of applications, including surveillance~\cite{angry_crowd, 8578776, 10.1007/978-3-030-58577-8_20}, traffic analysis~\cite{10815954, orlova2025simplifying, 10.1109/IROS40897.2019.8967556}, and healthcare~\cite{gabriel2024continuouspatientmonitoringai, 10.1145/3462203.3475892, 9684388}. While traditional VAD methods have shown strong performance, they often lack interpretability, failing to provide explanations for their predictions. This limitation reduces their trustworthiness in safety-critical real-world applications. To address this, recent studies have explored the integration of Large Language Models (LLMs) and Multimodal Large Language Models (MLLMs) to generate human-interpretable explanations alongside traditional VAD results, transitioning to a more holistic paradigm. In Video Anomaly Understanding (VAU), models not only detect anomalies but also describe them and reason about their cause, effects, and context.

However, existing VAU methods typically depend on fine-tuned MLLMs, which require costly annotations and complex training pipelines. A growing number of studies have begun exploring the zero-shot capabilities of these foundation models, leveraging techniques such as adapter tuning, video captioning, and prompt optimization. While these methods significantly reduce training overhead, they incur high inference costs due to dense and overlapping video processing, and the reliance on external auxiliary modules, such as video captioners or vision encoders.

In this work, we introduce PrismVAU, a streamlined approach to VAU that eliminates the need for fine-tuning MLLMs, using external encoders, or video captioning, significantly reducing inference overhead and enabling real-time accurate anomaly understanding. PrismVAU leverages a single off-the-shelf MLLM, VideoLLaMA-3~\cite{damonlpsg2025videollama3}, to perform anomaly scoring, explanation generation, and prompt optimization. It operates in two complementary stages: (1) a coarse anomaly scoring module, which estimates frame-wise anomaly scores based on the similarity between video frames and optimized textual anchors; and (2) an MLLM-based refinement module, which contextualizes and refines these initial scores using the corresponding video along with optimized system and user prompts. Both textual anchors and prompts are optimized using a weakly supervised Automatic Prompt Engineering (APE) framework, where the MLLM itself acts as the optimizer, requiring only video-level labels.



\section{Related Work}
\label{sec:relatedwork}
\subsection{Video Anomaly Detection} 
VAD aims to localize frames that contain abnormal events in a given video. Early VAD methods often approached the task as a one-class classification problem~\cite{liu2018ano_pred, tur2023unsupervised, 9710416, 9578232, 10204519, 10377180}, using datasets with only normal training samples like ShanghaiTech Campus~\cite{liu2018ano_pred} and CUHK Avenue~\cite{6751449}, among others~\cite{5539872, 9093457, 10203814}. These models learn to represent normal patterns via self-supervised pretext tasks, assuming anomalous input data would lead to poorer performance. 

Other methods have explored VAD in a weakly supervised setting~\cite{10203771, 8578776, 10204494, Li_Liu_Jiao_2022, 9710957, 9369126}, using datasets that provide 
only video-level labels during training, such as UCF-Crime~\cite{8578776}, XD-Violence~\cite{10.1007/978-3-030-58577-8_20}, MSAD~\cite{NEURIPS2024_a3c5af1f}, and TAD~\cite{10815954}. These models typically employ multiple instance learning frameworks and deliver superior performance compared to unsupervised one-class methods. 

With advancements in vision language models like CLIP~\cite{radford2021learningtransferablevisualmodels}, several VAD methods began to leverage the zero-shot capabilities of these models as either core components or auxiliary modules~\cite{wu2023vadclipadaptingvisionlanguagemodels, 10654921, 10667004, 10658364, 10657732}. These approaches typically integrate prompt tuning techniques, from learnable soft prompts to concept-based hard prompts, to steer the model’s representation space and enhance anomaly detection and categorization.

\subsection{Video Anomaly Understanding}
Despite these advances in VAD, traditional models primarily output anomaly scores without providing the rationale behind them, limiting their reliability in high-stakes, real-world scenarios. To overcome this, new studies have introduced a more holistic paradigm, VAU, integrating MLLMs and LLMs to produce human-understandable explanations alongside conventional VAD outputs. Among the different existing VAU methods, we distinguish between those that fine-tune MLLM components using instruction tuning~\cite{atang2024hawk, CUVA, du2024exploring, holmesvad, holmesvau, vadllama} and those that don't~\cite{vadllama, anomaly_ruler, lavad, vadsk, vera}.

On the one hand, methods that update MLLM weights are trained on multi-staged pipelines that demand extensive, fine-grained multi-modal annotations (\eg, anomaly types, timestamps, and natural language explanations), which are prohibitively  expensive to obtain. Moreover, they typically introduce custom LLM-based metrics and omit evaluation on standard VAD benchmarks. 

On the other hand, non-fine-tuned MLLM methods like AnomalyRuler~\cite{anomaly_ruler} and LAVAD~\cite{lavad} adopt a captioning-based strategy, where an MLLM performs dense captioning, and an LLM processes these captions to produce anomaly scores and contextual explanations, requiring multiple forward passes through both models. Alternatively, VERA~\cite{vera} learns guiding questions that are used to prompt the MLLM during inference to obtain coarse anomaly scores, which are further refined by aggregating scores from similar video segments and applying temporal smoothing and interpolation. While these methods generally benefit from a very efficient training phase compared to fine-tuning MLLMs, they incur high computational costs during inference due to their reliance on external LLMs, vision and text encoders, and processing videos in small, dense, and overlapping segments.

\subsection{Automatic Prompt Engineering}
Unlike traditional manual prompt engineering, which heavily relies on expert knowledge and static few-shot or Chain-of-Though approaches~\cite{CoT, fewshot}, APE methods like OPRO~\cite{opro} and Promptbreeder~\cite{fernando2023promptbreederselfreferentialselfimprovementprompt} discover effective prompt variants via systematic evolutionary strategies. In these frameworks, LLMs are leveraged as optimizers, where the optimization task is described in natural language. At each optimization step, the LLM generates new solutions using a meta-prompt that includes previously generated solutions and their performance scores. The top-performing prompts are then incorporated into the next iteration. This process not only enhances prompt effectiveness but also enables a more faithful evaluation of an LLM’s capabilities on a given task, eliminating the ambiguity of whether a poor result stems from the model or from suboptimal prompting. APE techniques have been successfully applied across multiple domains, such as natural language processing~\cite{opro_nlp, zhou2022large, choudhari2025promptsmartpayless}, arithmetic~\cite{fernando2023promptbreederselfreferentialselfimprovementprompt}, image generation~\cite{opro_img_gen}, and code generation~\cite{code_generation}. In contrast to soft prompt tuning, where a continuous vector is optimized in the embedding space, APE produces natural language prompts that are fully human-readable, easily modifiable, and transferable across models.

\section{Methodology}
\label{sec:methodology}
\begin{figure*}[t]
  \centering
   \includegraphics[width=0.9\linewidth]{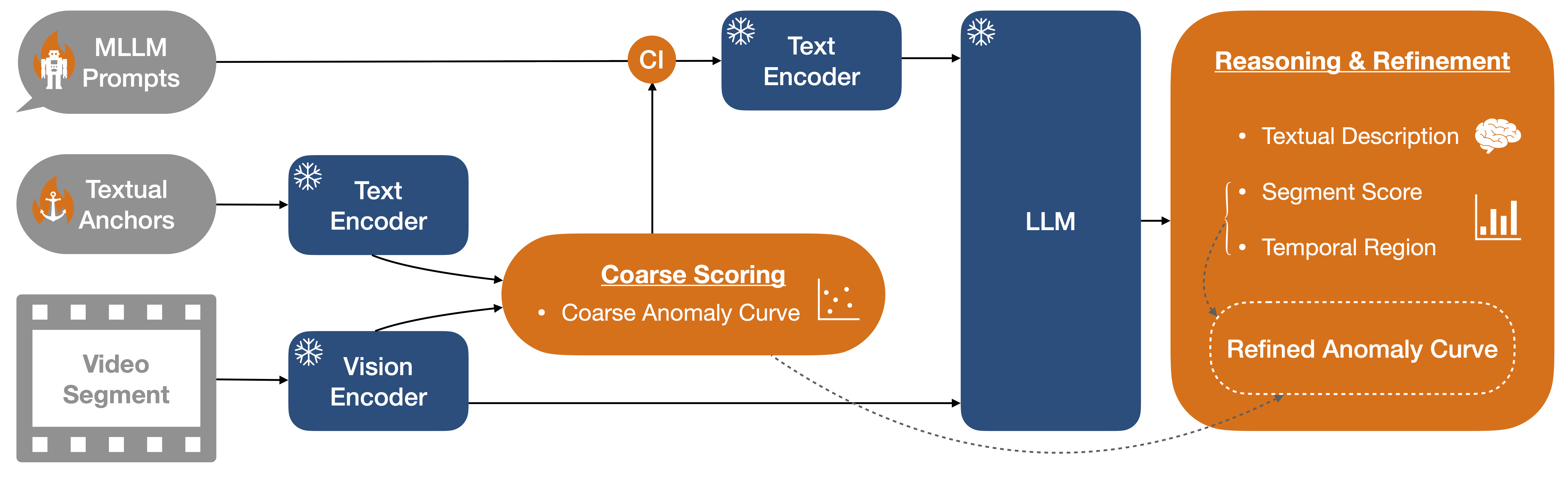}
   \caption{PrismVAU overview. Gray denotes inputs, blue indicates off-the-shelf MLLM modules, and orange highlights our contributions. A coarse anomaly curve is computed from the video segment and textual anchor embeddings, then injected (``CI’’) into the MLLM prompt. The MLLM outputs a textual explanation, a segment-level anomaly score, and an anomalous temporal region, which are used to refine the initial curve into a temporally precise and semantically informed anomaly signal.}
   \label{fig:prismvau_diagram}
\end{figure*}
PrismVAU, illustrated in~\cref{fig:prismvau_diagram}, is composed of two main modules for performing video anomaly understanding: a coarse anomaly scoring module and an MLLM-based reasoning and refinement module. The pipeline begins by processing video data in non-overlapping segments of up to 30 seconds, ensuring that each segment is likely to fully encompass a complete event, whereas shorter clips may fail to provide the temporal context needed for reliable anomaly understanding in a single forward pass. In the first stage, coarse anomaly scores are computed by measuring the similarity between embeddings of the video segment frames and a pair of optimized textual anchors. In the second stage, the MLLM is prompted with both the coarse predictions and the corresponding video segment to perform VAU directly, providing textual explanations and anomalous temporal regions. Finally, the MLLM outputs are integrated with the coarse anomaly curve to yield a temporally smoother and semantically richer anomaly signal. A more detailed description of each module is provided in the subsections below.

\subsection{Coarse Anomaly Scoring via Textual Anchors}
\label{subsec:alingment}
Given a video segment of duration $D$ seconds, frames are sampled at a rate of $Y$ frames per second (FPS), resulting in $N = \lfloor D \times Y \rfloor$ sampled frames. For each sampled frame $i$, its visual embedding $v_i$ is computed using the vision encoder of the MLLM, where the embedding is obtained by averaging the embeddings of the visual patches within the frame.

Let $t_{\text{norm}}$ and $t_{\text{abn}}$ denote the textual anchors representing generic descriptions of normal and abnormal scenes, respectively. Their corresponding text embeddings, $e_{\text{norm}}$ and $e_{\text{abn}}$, are obtained using the text encoder of the same MLLM by averaging the embeddings of the constituent text tokens.

The anomaly score $a_i$ of each sampled frame is computed using a softmax function over the cosine similarities between the visual embedding and the normal/abnormal text embeddings, $s_i^{\text{norm}}$ and $s_i^{\text{abn}}$:
\begin{equation}
    \label{eq:anomaly_score}
    a_i = \frac{\exp(s_i^{\text{abn}} / \tau)}{\exp(s_i^{\text{norm}} / \tau) + \exp(s_i^{\text{abn}} / \tau)} \, ,
\end{equation}
where $\tau$ is a temperature parameter that controls the sharpness of the softmax distribution, fixed to 0.07 following common CLIP usage for zero-shot classification~\cite{radford2021learningtransferablevisualmodels, chen2024understanding}. This results in an initial coarse and sparse anomaly score sequence $\{a_i\}_{i=1}^{N}$ that reflects the degree to which each sampled frame aligns with abnormal rather than normal semantics.\\

\begin{figure}[t]
  \centering
   \includegraphics[width=0.9\linewidth]{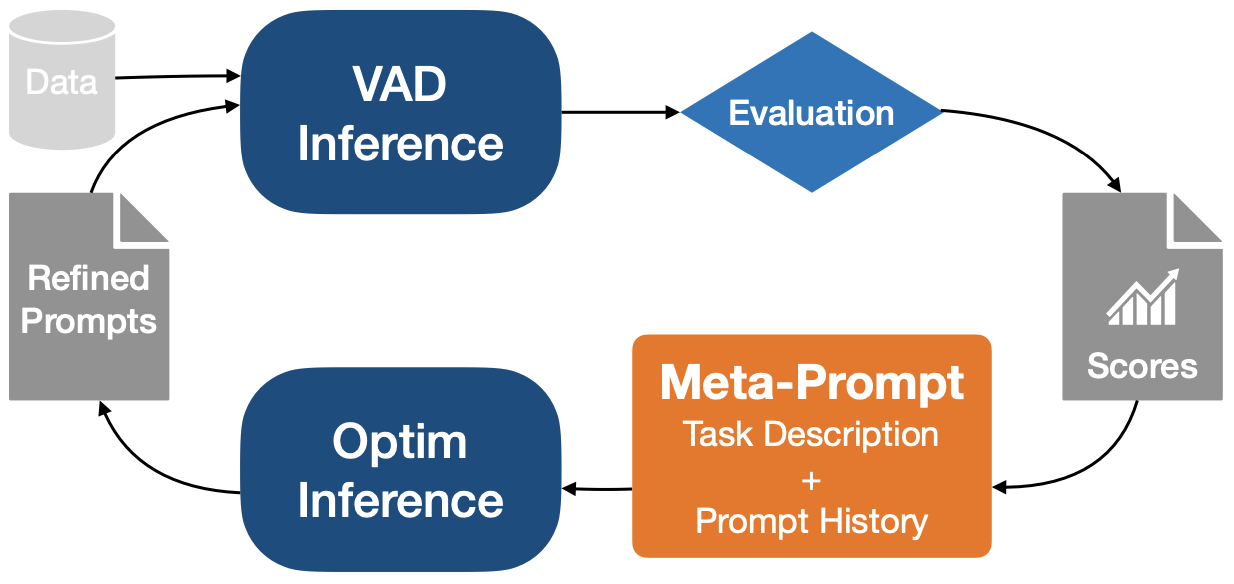}
   \caption{Diagram of the proposed APE framework for optimizing both textual anchors and MLLM prompts. The same MLLM is used for inference and optimization.}
   \label{fig:ape_diagram}
\end{figure}

The APE framework, depicted in~\cref{fig:ape_diagram}, is used to enhance the discriminability of the initial textual anchors (see~\cref{prompt:ta_base}). Following best practices from OPRO~\cite{opro}, a meta-prompt is constructed including the task description, the desired output format, a history of top-performing solutions, and a task reminder (see Supp. Material, Sec. A). The MLLM also serves as an optimizer to generate new prompt solutions. Each solution is then used to produce the corresponding set of sparse, per-frame anomaly scores across a collection of videos, $V$. Since only video-level labels are available during training, these scores are aggregated at the video level. To encourage diversity during optimization, multiple percentile functions are applied (\eg, 90th, 95th, and 99th percentiles), rather than relying solely on the maximum. The resulting video-level scores, $\{\hat{a_j}\}_{j=1}^{V}$, are evaluated by computing the Area Under the Curve (AUC) of the video-level Receiver Operating Characteristic (ROC) curve. The solution that achieves the highest score is reported for the next optimization step only if it is among the globally top-performing solutions.

This process ultimately yields an optimized pair of textual prompts, $t_{\text{norm}}^*$ and $t_{\text{abn}}^*$, which have been inferred implicitly from the dataset distribution and not manually defined. Rather than relying on human-written definitions of what constitutes a ``normal'' or ``abnormal'' scene, the APE framework encourages the model to distill these concepts directly from patterns in the video data. As a result, the learned prompts capture more nuanced and data-aligned representations of normality and abnormality, leading to improved generalization and performance across diverse video samples. This data-driven approach is particularly useful in real-world applications, where anomalous scenes may be complex, rare, or difficult to describe comprehensively.

\subsection{MLLM-Based Reasoning and Refinement}
While the coarse anomaly scores provide a rough estimate of the frame-level anomaly likelihood, they lack high-level temporal reasoning, contextual awareness, and semantic interpretability. To address these limitations, a second-stage refinement module is introduced, leveraging the capabilities of the MLLM to perform video anomaly understanding on individual video segments.

The model receives the video segment and is prompted to (i) determine whether an anomaly is present or not, (ii) localize it temporally by predicting start and end timestamps $u_s$ and $u_e$, (iii) assign an anomaly score $b\in[0,1]$ to the entire video segment, and (iv) generate a short textual explanation that links visual evidence to its prediction. Sparse frame-level scores are then derived by uniformly assigning the segment score to all sampled frames within the localized time interval, while frames outside receive a background value of 0, producing a step-like anomaly curve:
\begin{equation}
    \label{eq:mllm_anomaly_score}
    f(u) = 
    \begin{cases}
    b, & u_s \le u \le u_e \\
    0, & \text{otherwise}
    \end{cases}\, .
\end{equation}
Both the system and user prompts employed for the task are optimized using the same APE framework employed in the coarse stage. The initial and optimized prompts, as well as the meta-prompt template, are provided in Supp. Material, Sec. A.\\

Furthermore, we augmented the MLLM's input with the coarse anomaly predictions as additional guidance, motivated by the hypothesis that explicit similarity priors could help refine the model’s reasoning. Specifically, encoder-derived segment-level anomaly scores and associated statistics were inserted as textual annotations in the user prompt alongside the corresponding video segment. This design allows us to test whether the MLLM benefits from explicit conditioning on low-level similarity signals or whether such cues are already captured implicitly by its multimodal reasoning. Results in Subsection~\ref{subsec:ablations} confirm that supplying such explicit information leads to improved MLLM performance.\\

Lastly, to obtain a refined anomaly curve that is both semantically consistent with the MLLM’s judgments and temporally smooth, both coarse and MLLM sparse signals are processed using a Gaussian filter. The effective temporal smoothing window spans approximately $\pm2.5\sigma$ seconds at 1 FPS, capturing 99\% of the kernel's area. For the coarse anomaly predictions, $\sigma=2$ ($\approx5$s effective window) is manually selected, suppressing high-frequency noise while preserving short-duration anomalies and maintaining stable plateaus for longer events. This decision explicitly addresses the fact that coarse scores are computed independently for each sampled frame without temporal context. For the MLLM’s step-like predictions, a larger $\sigma=4$ ($\approx10$s effective window) is used to smooth abrupt transitions and enforce semantic consistency across adjacent segments. The resulting smoothed coarse and MLLM signals are combined through simple averaging. Finally, to match the original frame rate of the video, a parameter-free Fourier-based upsampling method is applied, preserving frequency content and avoiding distortions or temporal lag associated with naive frame-wise fill-forward approaches.
\section{Experiments and Results}
\label{sec:experiments}
\subsection{Datasets and Metrics}
We conduct experiments on two large-scale VAD datasets. Our main benchmark is UCF-Crime~\cite{8578776}, which consists of 1,900 untrimmed surveillance videos (128 hours) spanning 13 anomaly categories (\eg, robbery, fighting, vandalism) alongside normal events. UCF-Crime is typically evaluated using the frame-level ROC AUC, which measures how well anomalous and normal frames are separated. To further validate the generality of our approach, we also evaluate PrismVAU on XD-Violence~\cite{10.1007/978-3-030-58577-8_20}, a larger dataset with nearly 5k videos (217 hours) and 6 types of violent acts. XD-Violence is commonly evaluated using the frame-level Average Precision (AP), \ie, the area under the frame-level precision-recall curve, which focuses on measuring the ability to identify anomalous frames.

While ROC AUC and AP are threshold-free metrics suited for smooth anomaly curves, they are less appropriate for the MLLM outputs, which are step-like by design. These standard metrics can overestimate performance by emphasizing the ranking of anomalous versus normal frames rather than the precise temporal localization of anomalies. Therefore, the frame-level F1 score is also monitored using a fixed threshold of 0.5, allowing us to directly evaluate how well predicted segments cover true anomalous frames in a precise and interpretable way that balances missed detections and false alarms.

\subsection{Implementation Details}
VideoLLaMA-3-7B~\cite{damonlpsg2025videollama3} has been selected as the MLLM due to its state-of-the-art performance in video-language tasks such as long-video understanding and temporal reasoning. Moreover, it employs a pruning mechanism that removes temporally consecutive and visually similar tokens, a common characteristic of static surveillance footage, thereby improving computational efficiency.

Textual anchors are optimized using the APE framework for 100 iterations with a population size of 10, while retaining the top 10 pairs from previous iterations in the meta-prompt. Training is conducted on the full training set sampled at 1 FPS, with vision embeddings cached during the first iteration to accelerate subsequent evaluations.

Unlike anchor optimization, MLLM prompt evaluation requires a full forward pass through the entire model rather than just the encoders, making it considerably more expensive. To mitigate this, optimization is performed on a reduced subset of training videos sampled at 0.5 FPS, with a population size of 3 over 50 iterations, while retaining only the top 5 previous pairs. To ensure balanced coverage of the anomaly space under this computational budget, we curate a compact evaluation set with 3 randomly selected videos per anomaly type and an equal number of normal videos, totaling 78 clips.

At test time, all components are evaluated at 1 FPS. The MLLM employs greedy decoding to guarantee determinism and reproducibility, while training uses stochastic sampling. All optimization experiments fit within a single NVIDIA RTX 3090 GPU and complete in under 12 hours. On the same hardware, PrismVAU inference runs in real time, with an average latency of 0.13 s per processed frame and an effective throughput of 241 FPS under the 1 FPS sampling schedule.

\subsection{Comparison to State-of-the-art Methods}
A quantitative comparison with existing VAU and state-of-the-art (SOTA) VAD methods on UCF-Crime and XD-Violence datasets is reported in \cref{tab:sota}. Non-explainable approaches show steady gains with the adoption of weak supervision and CLIP-based designs. Fine-tuned VAU methods achieve the best overall results, underscoring the benefits of instruction tuning. Meanwhile, non-fine-tuned VAU approaches remain highly competitive, offering notable efficiency advantages alongside strong performance.

On both datasets, PrismVAU achieves the highest performance among non-fine-tuned VAU methods. On UCF-Crime, it reaches a ROC AUC of 87.82\% and an F1 score of 41.52\%, which is competitive with instruction-tuned VAU models and surpasses all traditional VAD methods except VadCLIP. On XD-Violence, PrismVAU obtains an AP of 78.68\% and an F1 score of 70.23\%, further demonstrating its robustness across datasets. Notably, on both benchmarks the frame-level false positive rate remains below 9\% when applying the intuitive 0.5 threshold, indicating that PrismVAU delivers reliable anomaly predictions without sacrificing precision, which makes it a practical and trustworthy approach for real-world applications.
\begin{table}
  \centering
  {\small{
  \begin{tabular}{@{}l|c|c@{}}
    Method  & UCF (ROC AUC) & XD (AP)\\
    \midrule
    \multicolumn{3}{c}{\textit{Non-explainable VAD Methods}} \\
    Tur \etal~\cite{tur2023unsupervised} & 66.85 & - \\
    FPDM~\cite{10377180} & 74.70 & - \\
    Sultani \etal~\cite{8578776} & 77.92 & 52.26 \\
    RTFM~\cite{9710957} & 84.30 & 77.81 \\
    MSL \cite{Li_Liu_Jiao_2022} & 85.62 & 78.59 \\
    Zhang \etal~\cite{10204494} & 86.22 & 81.43 \\
    UMIL \cite{10203771} & 86.75 & - \\
    OVVAD~\cite{10654921} & 86.40 & 66.53 \\
    PEL4VAD~\cite{10667004} & 86.76 & \underline{85.59} \\
    Chen \etal~\cite{10657732} & 86.83 & \textbf{88.21} \\ 
    TPWNG~\cite{10658364} & \underline{87.79} & 83.68 \\
    VadCLIP~\cite{wu2023vadclipadaptingvisionlanguagemodels} & \textbf{88.02} & 84.51 \\
    \midrule 
    \multicolumn{3}{c}{\textit{Fine-tuned VAU Methods}} \\
    VAD-LLaMA~\cite{vadllama} & 88.13 & -\\
    Holmes-VAU~\cite{holmesvau} & \underline{88.96} & \underline{87.68} \\  
    Holmes-VAD~\cite{holmesvad} & \textbf{89.51} & \textbf{90.67} \\
    \midrule
    \multicolumn{3}{c}{\textit{Non-fine-tuned VAU Methods}} \\
    LAVAD~\cite{lavad} & 80.28 & 62.01\\
    VERA~\cite{vera} & \underline{86.55} & \underline{70.54}\\
    \textbf{PrismVAU (ours)} & \textbf{87.82} & \textbf{78.68}\\
    \bottomrule
  \end{tabular}
  }}
  \caption{ROC AUC (\%) on UCF-Crime and AP (\%) on XD-Violence datasets of existing VAU and SOTA VAD methods. The best results per category are displayed in~\textbf{bold}, while the second best are \underline{underlined}.}
  \label{tab:sota}
\end{table}

\subsection{Ablation Studies}
\label{subsec:ablations}
In the following subsection, we present and analyze the intermediate results from both stages, coarse scoring via textual anchors and MLLM-based reasoning and refinement, and examine their twofold complementarity: (1) coarse predictions injected into MLLM prompts, and (2) coarse anomaly curve post-processing influenced by the MLLM outputs.\\

\noindent
\textbf{Textual anchors baseline.} As shown in~\cref{tab:baselines}, the APE framework effectively optimizes textual anchors. The initial solution performs below random chance at the frame level, with a ROC AUC of 41.63\%. After optimization, performance improves sharply, more than doubling both ROC AUC and F1 scores. Remarkably, this simple approach with textual anchors already achieves competitive results, reaching a ROC AUC above 84\% and surpassing several VAD and VAU methods.

These textual prompts evolve from short descriptions and enumerations of events into detailed scene-level narratives that emphasize environmental cues and human behaviors (see~\cref{prompt:ta}). This transition from abstract labels to visually grounded context indicates that MLLM encoders favor naturalistic language, leading to stronger alignment with image features. Moreover, optimized prompts provide distilled, generic descriptions that encapsulate the defining characteristics of anomalous and normal events in the UCF-Crime dataset.

\begin{figure}[t]
  \centering
  \begin{subfigure}{1\linewidth}
    \centering
    \includegraphics[width=0.95\linewidth]{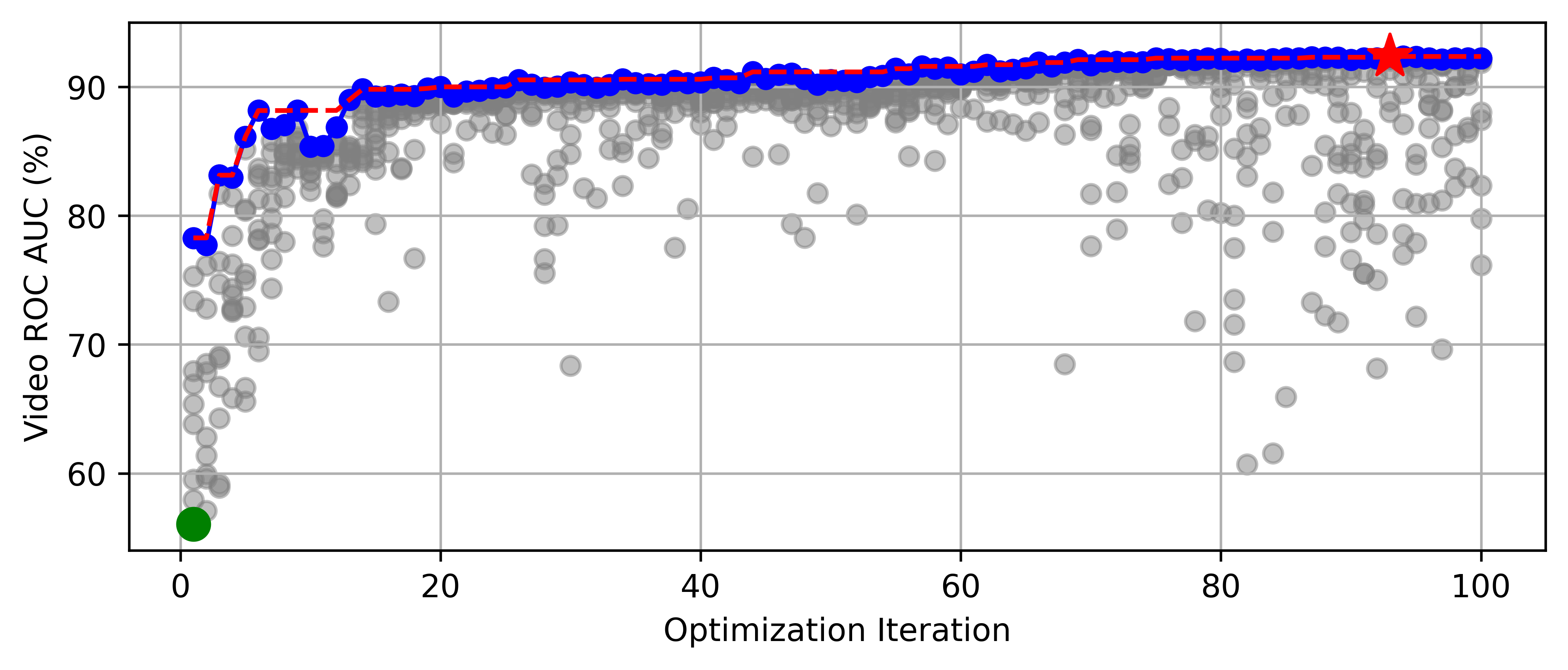}
    \caption{}
    \label{fig:anchors_curve}
  \end{subfigure}
  
  \vspace{0.5em}
  
  \begin{subfigure}{1\linewidth}
    \centering
    \includegraphics[width=0.95\linewidth]{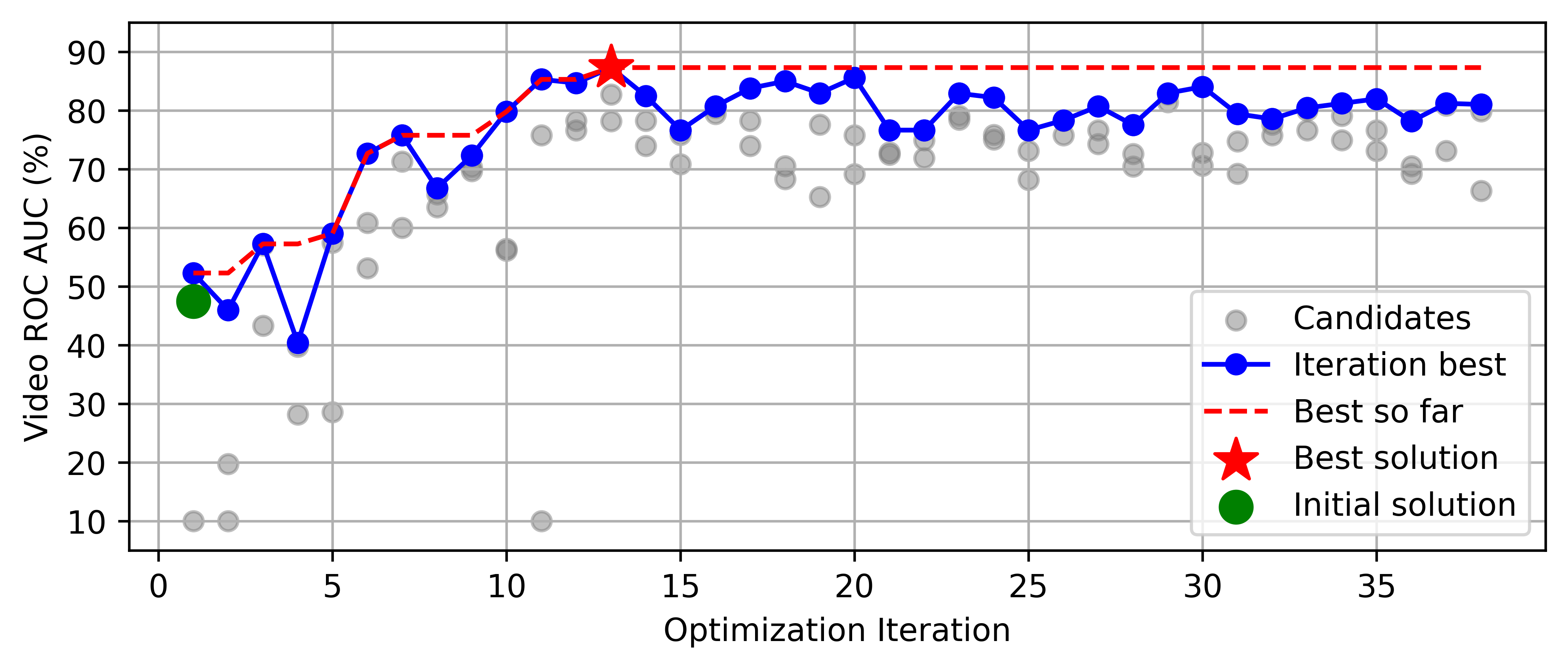}
    \caption{}
    \label{fig:mllm_curve}
  \end{subfigure}
  \caption{Learning curves produced by the APE optimization of textual anchors (a) and of the MLLM system and user prompts (b) on UCF-Crime.}
  \label{fig:ape_curves}
\end{figure}

The learning curve in~\cref{fig:anchors_curve} shows that early iterations yield rapid gains in video ROC AUC, indicating that the initial anchors were poorly aligned but quickly replaced with semantically stronger solutions. After roughly 20 iterations, the curve begins to plateau, reflecting the gradual refinement of existing top-performing prompts. The occasional presence of low-scoring candidates even in later iterations suggests that exploration is maintained throughout the process.
\begin{table}
  \centering
  {\small{
  \begin{tabular}{@{}l|c|c@{}}
    Method  & ROC AUC (\%) & F1 (\%)  \\
    \midrule
    TA & 41.63 & 13.11  \\
    TA$^*$ & 84.53 & 35.07  \\
    \midrule
    MLLM & 58.72 & 27.19 \\
    MLLM$^*$ & 70.78 & 39.45 \\
    \bottomrule
  \end{tabular}
  }}
  \caption{Baseline frame-level results on UCF-Crime. TA denotes the textual anchors baseline, MLLM refers to VideoLLaMA-3 without the injection of coarse priors, and $^*$ marks the use of optimized prompts. Both TA and MLLM predictions are upsampled using naive fill-forward.}
  \label{tab:baselines}
\end{table}
\begin{figure}[h]
    \centering
    \begin{subfigure}{0.95\linewidth}
        \centering
        \begin{tcolorbox}[colback=white, colframe=blue, boxrule=1pt, width=\linewidth]
            \begingroup
            \tiny\ttfamily
            \textbf{Normal Prompt:} A normal scene with usual activities.\\
            
            \textbf{Abnormal Prompt:} An abnormal scene with criminal activities such as stealing, robbery, burglary, abuse, road accidents, arrest, fighting, shoplifting, vandalism, assault, shooting, arson or explosion.
            
            \endgroup
        \end{tcolorbox}
        \caption{}
        \label{prompt:ta_base}
    \end{subfigure}
    
    \vspace{0.5em}
    
    \begin{subfigure}{0.95\linewidth}
        \centering
        \begin{tcolorbox}[colback=white, colframe=blue, boxrule=1pt, width=\linewidth]
            \begingroup
            \tiny\ttfamily        
            \textbf{Normal Prompt:} In a brightly lit, open public park or urban square, people are casually dressed in comfortable, everyday attire suitable for leisure activities. They stroll along well-maintained pathways lined with greenery, children play joyfully under the watchful eyes of guardians on lawns, and families engage in relaxed conversations while sitting on benches. The environment is characterized by clean sidewalks, residential homes with inviting exteriors, clear skies, and a peaceful ambiance, reflecting a sense of community life without any signs of distress or unusual activity.\\
            
            \textbf{Abnormal Prompt:} In a dimly-lit, cluttered urban setting, individuals exhibit aggressive body language indicative of illicit activities, such as confrontations or hiding in shadows. The environment is filled with debris, broken glass littering the ground, and objects suggestive of crime like weapons or stolen items. Emergency vehicles with flashing lights add urgency to the scene, contrasting sharply with the normal scene's tranquility. The presence of distress, aggression, or accidents stands out, creating an atmosphere charged with danger and chaos, where individuals' movements are quick and erratic, highlighting the risk of criminal activity.
            
            \endgroup
        \end{tcolorbox}
        \caption{}
        \label{prompt:ta_optim}
    \end{subfigure}
    \caption{Comparison of initial normal/abnormal textual anchors (a) with their optimized counterparts (b).}
    \label{prompt:ta}
\end{figure}

\noindent
\textbf{MLLM baseline.} As shown in~\cref{tab:baselines}, the APE framework is also effective at optimizing the MLLM system and user prompts for VAU, achieving an absolute improvement of over 10\% in both ROC AUC and F1 scores.

The initial prompt pair is high-level and generalized, instructing the model to simply detect anomalies and provide a score and description. In contrast, the optimized prompts adopt a more systematic and detailed approach, providing explicit guidance by listing specific behaviors to analyze, such as motion trajectories, object interactions, and contextual mismatches, thereby promoting temporally grounded reasoning (see Supp. Material, Sec. A). In contrast to the base solution, the specified output format is reinforced in both the system and user prompts. Overall, these refinements yield richer, context-aware anomaly detection and contribute to the observed performance gains.

\Cref{fig:mllm_curve} depicts the corresponding learning curve of the optimization process, which starts from a near-random solution. The curve demonstrates rapid learning, with the best solution found in less than 15 iterations, reaching a video-level ROC AUC of nearly 90\%. After this point, the performance plateaus, indicating that the optimization has converged to a highly effective prompt and early stopping is triggered.\\

\noindent
\textbf{Textual anchors vs MLLM.} \Cref{tab:baselines} highlights a key trade-off between the two approaches. Textual anchors produce smooth anomaly curves that rank anomalous frames more reliably, yielding superior threshold-free performance as measured by ROC AUC. In contrast, the MLLM generates step-like predictions that more accurately capture the temporal extent of anomalies, which improves segment-level detection performance at a fixed threshold, as reflected in the F1 score. The Pearson correlation between TA and MLLM predicted probabilities is moderate ($r=0.37$), while their error correlation is lower ($r=0.28$), indicating that the models often agree in general prediction trends but make mistakes on different instances. These results illustrate the complementary strengths not only of the evaluation metrics but, more importantly, of the methods themselves: textual anchors excel in global discriminability, while MLLM provides superior temporal precision and detailed textual reasoning, motivating their integration in a unified anomaly understanding system, PrismVAU. Baseline results on the XD-Violence dataset further corroborate this complementary behavior (see Supp. Material, Sec. B).

Building on this, we also investigate how well these components transfer across datasets focused on criminal activity. A new pair of textual anchors was distilled for XD-Violence, as the notion of abnormality slightly differs between datasets. In contrast, the MLLM prompts were reused directly from UCF-Crime, demonstrating their cross-dataset generalizability in detecting anomalies through motion trajectories, object interactions, and contextual mismatches.\\

\noindent
\textbf{Injection of coarse predictions.} Once optimal MLLM prompts for the VAU task were obtained, we explored the injection of coarse predictions as an additional input to the MLLM. This approach is motivated by the hypothesis that explicit similarity priors could complement the base MLLM performance. At the same time, it aligns the MLLM reasoning with the coarse anomaly curve.

We evaluated the effect of injecting coarse predictions into the MLLM input at varying levels of granularity, all provided in JSON format. The injected information includes the coarse video label (indicating whether the textual anchors module detected an anomaly in the video segment), the coarse segment-level abnormality score, and the start and end timestamps of the most anomalous coarse region, if any. Additionally, we incorporated frame-level statistics derived from coarse scores, such as the maximum score and its timestamp, the median, the mean, the standard deviation, and the total time duration exceeding the 0.5 anomaly threshold. As shown in~\cref{tab:coarse_injection}, progressively providing more detailed coarse information consistently improves ROC AUC, with the full set of coarse features achieving the highest performance.\\
\begin{table}
  \centering
  {\small{
  \begin{tabular}{@{}c|c|c|c|c@{}}
    L & S & T & F & ROC AUC (\%)\\
    \midrule
               &            &               &               & 70.78 \\
    \checkmark &            &               &               & 72.91 \\
    \checkmark & \checkmark &               &               & 73.68 \\
    \checkmark & \checkmark & \checkmark    &               & 74.18 \\
    \checkmark & \checkmark & \checkmark    & \checkmark    & 76.97 \\
    \bottomrule
  \end{tabular}
  }}
  \caption{Results of progressively injecting coarse predictions into the MLLM input on UCF-Crime. ``L'' denotes video label, ``S'' video score, ``T'' temporal region, and ``F'' frame statistics.}
  \label{tab:coarse_injection}
\end{table}

\noindent
\textbf{Per-category performance.} PrismVAU surpasses both its components across most anomaly categories, achieving its strongest performance in \textit{assault}. TA is most effective when anomalies exhibit distinctive visual cues, while MLLM excels in cases requiring contextual, relational, or subtle motion understanding. A more detailed per-category analysis is provided in Supp. Material, Sec. C.

\subsection{Qualitative Results}
\begin{figure*}[t]
  \centering
   \includegraphics[width=0.95\linewidth]{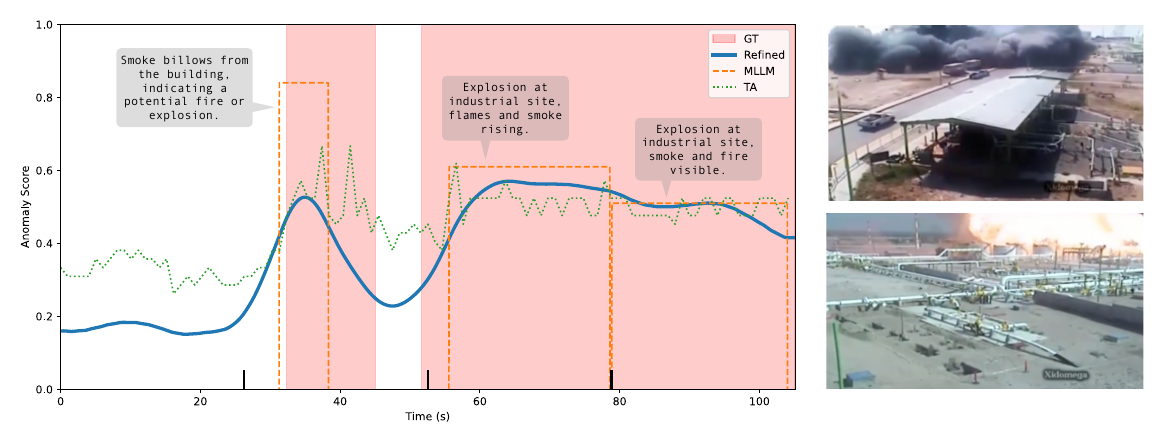}
   \caption{Results on \textit{Explosion033\_x264} from UCF-Crime. Ground-truth (GT) anomalous intervals are compared against the coarse anomaly curve from the textual anchors (TA), the step-like MLLM predictions, and the refined anomaly curve. Human-readable explanations generated by the MLLM are shown as speech bubbles. Bold ticks on the x-axis indicate video segment partitions used when processing the untrimmed video. Sample frames on the right illustrate key anomaly events.}
   \label{fig:anomaly_curve}
\end{figure*}

The coarse anomalous curves derived from the textual anchors exhibit high-frequency noise but still capture the overall anomaly trend. The MLLM, in turn, produces step-like predictions that align well with ground-truth boundaries in most cases and minor offsets can be partly attributed to annotation ambiguity, as it is not always clear when certain behaviors begin or end (\eg, lingering near items may precede the actual theft). Occasionally, the MLLM detects subtle unannotated anomalies that plausibly precede or follow the main event. PrismVAU's refined curve achieves sharp temporal localization while also improving frame-level discriminability. 

PrismVAU's generated textual explanations are representative of the observed events, though confusions may arise between closely related anomaly categories (\eg, abuse, assault, fight, and robbery), which often co-occur within the same event. In some cases, descriptions are enriched by leveraging textual cues in the video, such as captions or visible signage. To further assess explanation quality, semantic similarity is computed between the generated descriptions and the HIVAU-70k~\cite{holmesvau} ground-truth annotations using MPNet~\cite{NEURIPS2020_c3a690be} embeddings. An average similarity of 0.56 is obtained, indicating that PrismVAU generally captures the core semantics of the underlying anomalous events. Additional details of this procedure are provided in Supp. Material Sec. D.

\Cref{fig:anomaly_curve} showcases PrismVAU’s anomaly detection results on a test video from UCF-Crime, where it successfully identifies and describes two explosion events. Additional case studies on videos from both benchmarks are presented in Supp. Material, Sec. E, further demonstrating PrismVAU’s ability not only to detect anomalies but also to contextualize them with semantically meaningful explanations.

Another valuable feature of PrismVAU is its interactive chat interface, as no weight fine-tuning is applied. Users can engage with the model to audit system errors, elaborate on correct predictions, and explore counterfactual or hypothetical scenarios, as shown in~\cref{fig:mllm_interaction_1} and in Supp. Material, Sec. E. This interactive setup improves transparency and interpretability while also demonstrating the model’s ability to reason about cause, intent, and context, essential capabilities for trustworthy video anomaly detection.
\begin{figure}
  \centering
   \begin{tcolorbox}[colback=white, colframe=orange, boxrule=1pt, width=\linewidth]
        \begingroup
        \tiny\ttfamily
        \textbf{PrismVAU:} \{"abnormality": 0.00, "description": "No visible criminal behavior detected in the video footage."\}\\
        
        \textbf{User:} Describe the scene and explain why it is considered normal.\\
        
        \textbf{PrismVAU:} The video shows a typical city street scene with parked cars and a pedestrian walking by. The presence of parked cars indicates it is likely a residential or commercial area. The pedestrian's movement is normal for a city environment. No unusual crowd clustering, suspicious objects, or contextual mismatches are observed, suggesting that the scene does not indicate any criminal activity.\\
        
        \textbf{User:} What would happen if a man started breaking the windows of parked cars?\\
        
        \textbf{PrismVAU:} If a man started breaking the windows of parked cars, it would be considered criminal behavior. This action would fall under vandalism and would likely be flagged as an anomaly by the system due to its unusual nature and the destruction of property.\\
        
        \endgroup
    \end{tcolorbox}

   \caption{Interactive dialogue with PrismVAU on a normal scene, \textit{Normal\_Videos\_877\_x264} from UCF-Crime.}
   \label{fig:mllm_interaction_1}
\end{figure}

\section{Summary and Future Work}
\label{sec:conclusions}
We introduced PrismVAU, a lightweight and effective approach to real-time VAU. Unlike prior methods that require fine-tuned MLLMs, auxiliary modules, or dense and overlapping video processing, PrismVAU operates efficiently with a single off-the-shelf MLLM. It combines a coarse anomaly scoring module based on textual anchors with an MLLM-driven reasoning and refinement stage that contextualizes anomalies via prompts and lightweight post-processing. Both anchors and prompts are optimized using a weakly supervised APE framework, with the MLLM itself serving as the optimizer. Experimental results on standard VAD benchmarks demonstrate that PrismVAU achieves competitive anomaly detection performance while also providing interpretable explanations of anomalous events.\\

While this work focuses on criminal behavior, PrismVAU can be extended to detect other types of events or actions by re-optimizing both anchors and prompts. Although we have evaluated PrismVAU using VideoLLaMA-3, the pipeline is model-agnostic and can be adapted to other MLLMs with minimal effort. Future work should also explore the development of standardized benchmarks and metrics for evaluating textual explanations, as current fine-tuned methods rely on custom datasets and ad hoc criteria. Establishing such standards would enable more robust and objective comparisons across methods, accelerating the development of transparent and reliable VAU systems.\\

\noindent
\textbf{Ethical considerations.} The definition of criminal actions in this work relies on annotations from existing, recognized datasets. We acknowledge that these annotations may be imperfect, context-dependent, and subject to sociocultural variation. Accordingly, PrismVAU is presented as a step toward ethical autonomous criminal activity detection and is not intended for real-world deployment without careful ethical, practical, and regulatory consideration.\\

\noindent
\textbf{Acknowledgements.} This work has been partially supported by the Spanish project PID2022-136436NB-I00, by ICREA under the ICREA Academia programme, and by the Milestone Research Program at the University of Barcelona.
{
    \small
    \bibliographystyle{ieeenat_fullname}
    \bibliography{main}
}

\end{document}


\maketitle  
\section{Prompts and Meta-prompts}

In this section, we provide the meta-prompt used as user prompt for optimizing the textual anchors (see~\cref{fig:meta_alignment}). The meta-prompt used for optimizing the MLLM prompts for the VAU task (see~\cref{fig:meta_vlm}). In both cases, the \textcolor{red}{red} text corresponds to the placeholders representing initial solutions and their associated scores. The initial and optimized MLLM prompts for VAU are also reported in~\cref{prompt:mllm}, where the \textcolor{blue}{blue} text corresponds to the portions manually augmented to incorporate coarse anomaly predictions as additional context in JSON format. Finally, the optimized textual anchors for the XD-Violence dataset are displayed in~\cref{prompt:ta_xd}.\\

\begin{figure*}[h]
  \centering
   \begin{tcolorbox}[colback=white, colframe=blue, boxrule=1pt, width=0.95\textwidth]
        \begingroup
        \tiny\ttfamily
        You are an expert in designing visually grounded prompts for vision-language models, specifically for binary image classification tasks. Your goal is to generate two detailed and distinct prompts, one for normal scenes and one for abnormal scenes, that are visually grounded in the specific physical characteristics of each category. These prompts will be used for zero-shot classification with CLIP, so it’s crucial that they clearly differentiate between typical, non-threatening actions and suspicious, abnormal behaviors.\\
        
        For the normal, scenes describe typical, non-threatening human behavior in common public or indoor settings. Include physical traits of people (\eg, posture, clothing), common environments (\eg, streets, parks, offices), and ordinary activities (\eg, walking, sitting, talking). For the abnormal scenes,  describe behavior that stands out as unusual, dangerous, or criminal. Include specific visual details about suspicious actions, the physical environment (\eg, restricted spaces, chaotic areas), objects that are part of the scene (\eg, weapons, stolen items), and the presence of distress, aggression, or accidents.\\

        You must provide the output in the following format:\\
        Normal Prompt: <Generated prompt for normal scenes>\\
        Abnormal Prompt: <Generated prompt for abnormal scenes>\\

        The following are examples of past prompts with their obtained classification scores. The score ranges from 0 to 100, where 100 is the best possible value. Use these examples and their scores as a reference to improve prompt construction.\\

        Example 1:\\
        - Normal Prompt: \textcolor{red}{\{initial normal prompt\}}\\
        - Abnormal Prompt: \textcolor{red}{\{initial abnormal prompt\}}\\
        - Score: \textcolor{red}{\{initial pair score\}}\\

        \textit{(... more top-performing previous examples, sorted in ascending order ...)}\\
        
        Your goal is to create a more effective prompt that enhances performance in distinguishing abnormal from normal scenes. Prioritize visually grounded concepts, ensuring the descriptions are rich and capture defining physical traits, with adjectives and objects typically present in each. Do not reference scores in your response, just focus on achieving the most effective differentiation.
        
        \endgroup
    \end{tcolorbox}

   \caption{Meta-prompt used for optimizing textual anchors.}
   \label{fig:meta_alignment}
\end{figure*}

\begin{figure*}[t]
  \centering
   \begin{tcolorbox}[colback=white, colframe=blue, boxrule=1pt, width=0.95\textwidth]
        \begingroup
        \tiny\ttfamily
        You are an expert prompt engineer specialized in designing prompts for Multimodal LLMs (Video‑LLMs) to perform temporal anomaly detection in urban surveillance. Your goal is to propose an improved System Prompt and User Prompt that guide the model to:\newline
        
        1. Analyze object dynamics, motion, interactions, and context over time.\newline
        2. Detect criminal behaviors (\eg, assault, arson, burglary, vandalism, shoplifting) by attending to subtle or compound indicators such as sudden acceleration, unusual crowd clustering, suspicious objects and contextual mismatches.\newline
        3. Identify and report the exact time interval of the anomaly (in seconds, precise to 2 decimal places).\newline
        4. Assign an abnormality score (0.00-1.00, rounded to two decimal places) that reflects the severity of the detected behavior.\newline
        5. Generate a concise description ($\leq$ 100 characters) summarizing what was observed and why it was flagged.\newline
        
        The MLLM should reason step-by-step before producing its final response. That is, it should:\newline
        - First, analyze motion patterns, object states, and interaction dynamics across time.\newline
        - Then, consider contextual clues to determine whether a behavior is anomalous.\newline
        - Finally, output the result in this specific JSON format:\newline
        \{\newline
        "start\_time": <float seconds, 2dp>,\newline
        "end\_time": <float seconds, 2dp>,\newline
        "abnormality": <float between 0.00 and 1.00, 2dp>,\newline
        "description": <concise summary, max 100 characters>\newline
        \}\newline
        If no anomaly is detected, respond with \{\}.\newline

        You will now see examples of previous prompt pairs with their obtained performance score, where 100 is the best possible value. Use their language, structure, and specificity as inspiration to improve prompt design.\newline

        Example 1:\newline
        - System Prompt: \textcolor{red}{\{initial system prompt\}}\newline
        - User Prompt: \textcolor{red}{\{initial user prompt\}}\newline
        - Score: \textcolor{red}{\{initial pair score\}}\newline

        \textit{(... more top-performing previous examples, sorted in ascending order ...)}\newline

        Task:\newline
        Generate a stronger System Prompt + User Prompt pair. Focus on:\newline
        - Encouraging temporal attention to motion patterns and interactions\newline
        - Emphasizing precise format instructions\newline
        - Being concise yet comprehensive\newline

        Return ONLY the following format, with no additional text, explanations, or scores:\newline
        System Prompt: <...>\newline
        User Prompt: <...>
        
        \endgroup
    \end{tcolorbox}

   \caption{Meta-prompt used for optimizing the VAU prompts of the MLLM.}
   \label{fig:meta_vlm}
\end{figure*}

\begin{figure*}[h]
    \centering
    \begin{subfigure}{0.95\linewidth}
        \centering
        \begin{tcolorbox}[colback=white, colframe=blue, boxrule=1pt, width=\linewidth]
            \begingroup
            \tiny\ttfamily
            \textbf{System Prompt:} You are a Video-LLM for urban surveillance anomaly detection. Your task is to analyze video footage to find anomalies and criminal behavior. If present, localize them temporally by predicting start and end timestamps, assign an anomaly score to the entire anomalous event, and generate a textual explanation that links visual evidence to its prediction.\\
        
            You must always return your results in the following JSON format:\\
            \{\\
            "start\_time": <float in seconds, precise to 2 decimal places>,\\
            "end\_time": <float in seconds, precise to 2 decimal places>,\\
            "abnormality": <float between 0.00 and 1.00, rounded to 2 decimal places - represents how severe the anomaly is>,\\
            "description": <string, concise summary of the anomaly, max 100 characters, detailing key observations and evidence>\\
            \}\\
            If no anomalies are found, you should return an empty JSON, \{\}.\\
            
            \textbf{User Prompt:} Analyze the following video and return only the required JSON.
            
            \endgroup
        \end{tcolorbox}
        \caption{}
        \label{prompt:mllm_base}
    \end{subfigure}
    
    \vspace{0.5em}
    
    \begin{subfigure}{0.95\linewidth}
        \centering
        \begin{tcolorbox}[colback=white, colframe=blue, boxrule=1pt, width=\linewidth]
            \begingroup
            \tiny\ttfamily
            \textbf{System Prompt:} As a Video Language Model specialized in urban surveillance, your primary mission is to meticulously analyze video content for temporal anomalies in object dynamics, motion patterns, interactions, and contextual clues over time. Prioritize identifying criminal behaviors such as assault, arson, burglary, vandalism, and shoplifting by focusing on subtle or compound indicators like sudden acceleration, unusual crowd clustering, suspicious objects, and contextual mismatches. Maintain a high level of temporal attention to motion patterns and interactions, ensuring precise frame-by-frame analysis to make accurate decisions. Assign an abnormality score between 0.00 and 1.00, rounded to two decimal places, based on the severity of the detected behavior. Provide a concise summary ($\leq$ 100 characters) summarizing what was observed and why it was flagged. If no anomaly is detected, return an empty JSON. \textcolor{blue}{Use the provided statistical summary as a preliminary signal but ensure your own thorough, explicit analysis is conducted.}\\

            Document your findings in the following JSON format, providing detailed reasoning and explicit documentation of all evidence used in each step:\\
            \{\\
            "start\_time": <float in seconds, precise to 2 decimal places>,\\
            "end\_time": <float in seconds, precise to 2 decimal places>,\\
            "abnormality": <float between 0.00 and 1.00, rounded to 2 decimal places - represents how severe the anomaly is>,\\
            "description": <string, concise summary of the anomaly, max 100 characters, detailing key observations and evidence>\\
            \}\\

            \textbf{User Prompt:} \textcolor{blue}{A preliminary system has provided the following statistical summary as a prior signal:\\
            \{\\
            "coarse\_video\_label": <string>,\\
            "coarse\_video\_abnormality\_score": <float>,\\
            "coarse\_temporal\_region": <list>,\\
            "coarse\_frame\_abnormality\_stats":\{\\
                \hspace*{2em}"max": <float>,\\
                \hspace*{2em}"time\_max": <float>,\\
                \hspace*{2em}"median": <float>,\\
                \hspace*{2em}"mean": <float>,\\
                \hspace*{2em}"std": <float>,\\
                \hspace*{2em}"abnormal\_seconds": <float>\\
                \hspace*{2em}\}\\
            \}\\
            }
            
            Carefully examine the provided video footage, ensuring that you adhere strictly to the guidelines outlined in the System Prompt. Focus on precise temporal attention to motion patterns and interactions, prioritizing frame-by-frame analysis while considering contextual clues to identify anomalous behaviors. Assign an abnormality score accurately and provide a concise summary as specified. If no anomaly is found, respond with an empty JSON.\\
            \{\\
            "start\_time": <float in seconds, precise to 2 decimal places>,\\
            "end\_time": <float in seconds, precise to 2 decimal places>,\\
            "abnormality": <float between 0.00 and 1.00, rounded to 2 decimal places - represents how severe the anomaly is>,\\
            "description": <string, concise summary of the anomaly, max 100 characters, detailing key observations and evidence>\\
            \}\\
            
            \endgroup
        \end{tcolorbox}
        \caption{}
        \label{prompt:mllm_optim}
    \end{subfigure}
    \caption{Comparison of initial system and user MLLM prompts for VAU (a) with their optimized counterparts (b).}
    \label{prompt:mllm}
\end{figure*}

\begin{figure*}[t]
  \centering
   \begin{tcolorbox}[colback=white, colframe=blue, boxrule=1pt, width=0.95\textwidth]
        \begingroup
        \tiny\ttfamily
        \textbf{Normal Prompt:} In a tranquil, well-lit suburban neighborhood during the daytime, individuals are casually dressed in comfortable attire suitable for leisure activities. They engage in routine behaviors such as walking dogs, children playing on swings and slides, and couples enjoying picnics under the shade of mature trees. The environment features clean pathways lined with blooming flowers, lush greenery, and clear blue skies overhead, reflecting a sense of calm and normalcy. Individuals display relaxed postures, indicating ease and contentment, with no signs of distress or urgency present in this peaceful community space.\\

        \textbf{Abnormal Prompt:} Amidst the shadowy corners of an urban alleyway at night, individuals exhibit secretive behavior with tense body language, suggesting potential illicit intent. The surroundings are cluttered with debris, graffiti, and restricted spaces, creating a tense atmosphere indicative of danger and criminal intent. People are dressed in dark clothing that blends into the shadows, adding to the sense of illicit activity. Signs of recent criminal incidents, such as litter, abandoned vehicles, and visible weapons, contribute to an overall sense of urgency and imminent threat. Visible objects associated with criminal activity further enhance the sense of danger and criminal intent, while individuals display signs of distress, aggression, or involvement in suspicious activities, amplifying the imminent threat and criminal intent in the scene.\\
        
        \endgroup
    \end{tcolorbox}

   \caption{Distilled textual anchors of the XD-Violence dataset.}
   \label{prompt:ta_xd}
\end{figure*}

\clearpage
\section{Results on XD-Violence}
In this section, we present the baseline results on the XD-Violence dataset. A new pair of textual anchors was distilled to account for dataset-specific notions of abnormality, achieving a substantial improvement in performance over unoptimized anchors. In contrast, the MLLM prompts from the UCF-Crime dataset were reused directly, demonstrating their cross-dataset generalizability in detecting anomalies through motion trajectories, object interactions, and contextual mismatches. The results in \cref{tab:baselines_xd} further highlight the complementary strengths of these methods; textual anchors achieve higher AP, reflecting stronger global discriminability, while the MLLM provides superior temporal precision, as reflected in the F1 scores.
\begin{table}[h]
  \centering
  {\small{
  \begin{tabular}{@{}l|c|c@{}}
    Method  & AP (\%) & F1 (\%)  \\
    \midrule
    TA & 42.15 & 39.03  \\
    TA$^*$ & 74.66 & 56.36  \\
    \midrule
    MLLM & 59.44 & 26.81 \\
    MLLM$^*$ & 65.07 & 59.98 \\
    \bottomrule
  \end{tabular}
  }}
  \caption{Baseline frame-level results on XD-Violence. TA denotes the textual anchors baseline, MLLM refers to VideoLLaMA-3 without coarse priors, and $^*$ marks the use of optimized prompts.}
  \label{tab:baselines_xd}
\end{table}
\clearpage

\section{Performance Across Anomaly Types}
\begin{figure*}[h]
  \centering
   \includegraphics[width=0.9\linewidth]{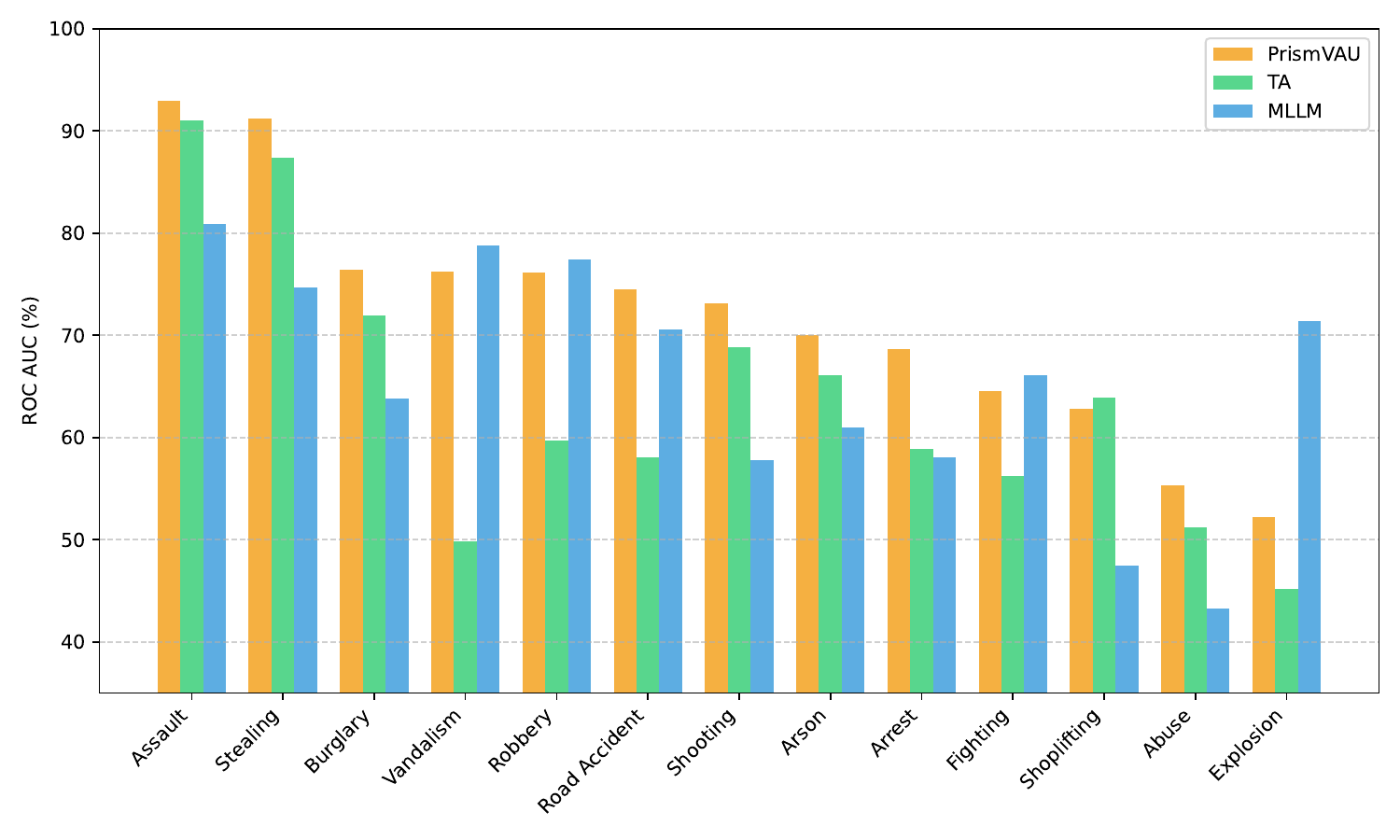}
   \caption{Per-category ROC AUC performance comparison of PrismVAU and its components on UCF-Crime.}
   \label{fig:categories}
\end{figure*}

As depicted in~\cref{fig:categories}, PrismVAU outperforms both its components in 8 of 13 categories. Owing to its joint design, PrismVAU can compensate when either TA or the MLLM underperforms, yielding its highest ROC AUC on \textit{assault} and lowest on \textit{explosion}. The largest performance gains are observed in \textit{vandalism} and \textit{road accident} relative to TA, and in \textit{stealing} relative to the MLLM.\\

Textual anchors tend to excel when anomalies exhibit distinctive visual patterns or strong visual-text alignment, particularly when static cues suffice, as in \textit{shooting} (weapon, pose, and scene cues). In contrast, the MLLM performs better when context, relational reasoning, or subtle temporal cues are critical, such as in \textit{vandalism} or \textit{robbery}.\\

The most challenging categories are \textit{explosion}, \textit{abuse}, and \textit{shoplifting}. Low performance on \textit{explosion} may result from missed temporal cues due to sparse sampling and single-frame embeddings. \textit{Abuse} likely suffers from high variability and ambiguous semantics. \textit{Shoplifting} may involve subtle hand-object interactions under occlusion, making fine-grained behavior recognition difficult.
\clearpage

\section{Semantic Similarity Analysis}
Directly evaluating PrismVAU’s generated explanations is challenging because UCF-Crime does not include textual annotations. To approximate their semantic alignment with the underlying anomalous events, we instead leverage the HIVAU-70k dataset. Built on top of UCF-Crime and XD-Violence, HIVAU-70k provides clip-, event-, and video-level free-text annotations that judge, describe, and analyze the presence or absence of anomalies.\\

Since the objective is to assess meaning rather than lexical overlap, an embedding-based similarity metric has been employed to evaluate explanation quality. Specifically, for each video $V_i$, textual embeddings are computed for all its predicted explanations $\{p_1, \ldots,p_{N}\}$ and ground truth sentences $\{g_1, \ldots,g_{M}\}$, after which a cosine similarity matrix is formed over all prediction-ground-truth pairs. Each prediction is assigned its maximum similarity to any ground-truth sentence, the video-level score is obtained by averaging these maxima, and the dataset-level score is computed as the mean across videos. MPnet (\textit{all-mpnet-base-v2}) has been used as the text encoder due to its superior semantic representation capabilities, which enable reliable assessment of conceptual alignment between predicted and reference explanations.\\

Cosine similarity values can be interpreted approximately as: 0.0–0.2 (unrelated), 0.2–0.4 (weakly related), 0.4–0.6 (moderately aligned), 0.6–0.8 (strongly aligned), and 0.8–1.0 (nearly identical). PrismVAU obtains an average score of 0.5558, indicating moderate semantic alignment between predicted anomaly descriptions and ground-truth text. This outcome is consistent with the inherent challenges introduced by: short (sub-100-character) anomaly-focused predictions versus context-rich annotations, mismatches in temporal granularity and abstraction, variation in domain-specific vocabulary, and the absence of textual supervision during training. Despite these factors, the score reflects that PrismVAU reliably captures the correct semantic category of anomalous events, despite differences in phrasing or level of detail. The metric therefore provides a principled quantitative estimate of explanation quality and complements the anomaly-detection evaluation by demonstrating that PrismVAU’s explanations are both plausible and semantically meaningful.
\clearpage

\section{Additional Qualitative Results}
In this section, we present additional qualitative results and case studies on both the UCF-Crime and XD-Violence datasets. Ground-truth (GT) anomalous intervals are compared against the coarse anomaly curve from the textual anchors (TA), the step-like MLLM predictions, and the refined anomaly curve. Human-readable explanations generated by the MLLM are shown as speech bubbles. Bold ticks on the x-axis indicate video segment partitions used when processing the untrimmed video. Sample frames on the right illustrate key anomaly events.\\

\begin{figure*}[h]
  \centering
   \includegraphics[width=0.95\linewidth]{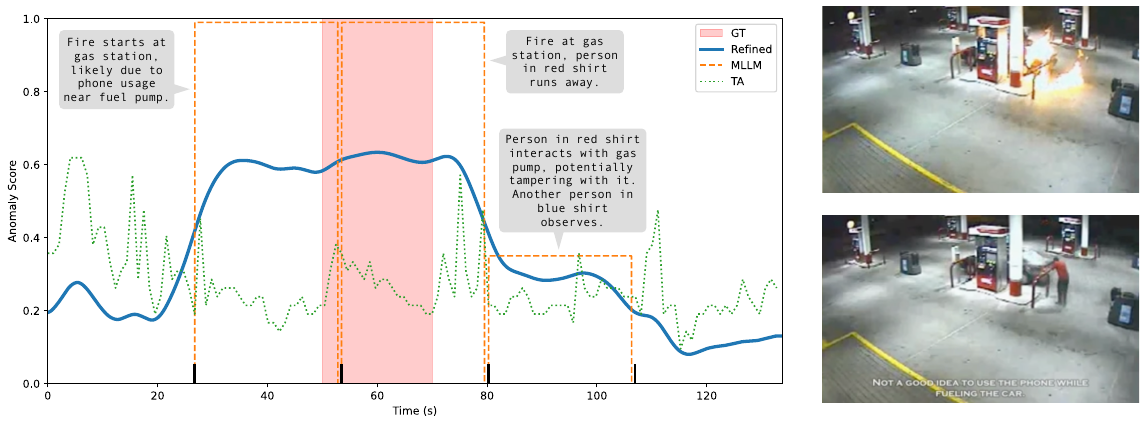}
   \caption{Results on \textit{Explosion002\_x264} from UCF-Crime.}
   \label{fig:Explosion002}
\end{figure*}

\begin{figure*}[h]
  \centering
   \includegraphics[width=0.97\linewidth]{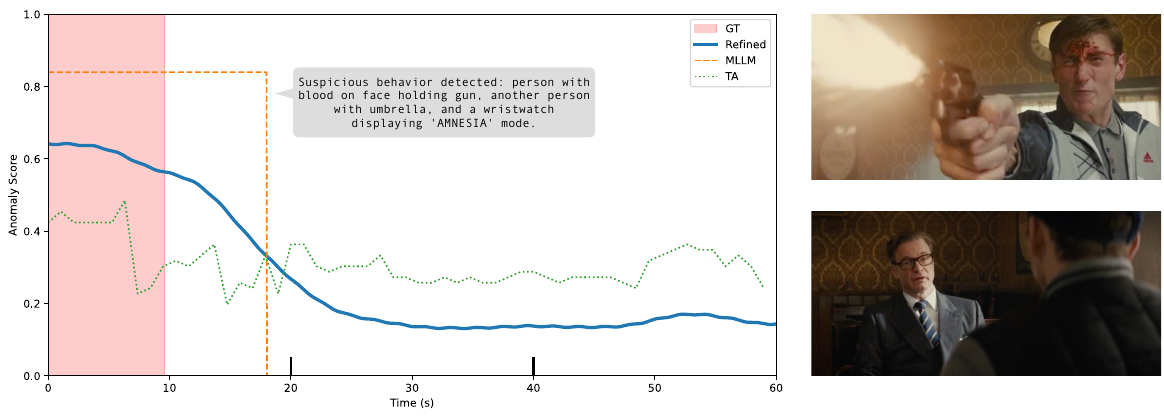}
   \caption{Results on \textit{Kingsman.The.Secret.Service.2014\_\_\#00-22-10\_00-23-10\_label\_B2-0-0} from XD-Violence.}
   \label{fig:xd2}
\end{figure*}

\begin{figure*}[h]
  \centering
   \includegraphics[width=0.95\linewidth]{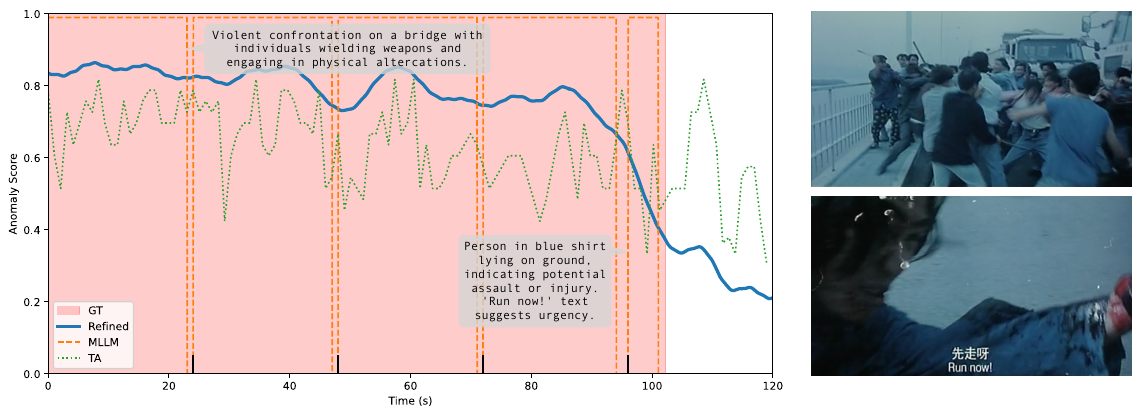}
   \caption{Results on \textit{Young.And.Dangerous.I.1996\_\_\#0-46-57\_0-48-57\_label\_B1-0-0} from XD-Violence.}
   \label{fig:xd3}
\end{figure*}

\Cref{fig:Explosion002} illustrates PrismVAU’s anomaly detection results on a UCF-Crime test video containing an explosion event. In this example, the coarse anomaly curve is noisy, while the MLLM predictions are both accurate and temporally anticipatory, preceding the ground-truth labels. The explanations demonstrate the MLLM’s ability to associate the explosion with textual clues suggesting it is phone-related. A similar capability to leverage captions and textual evidence for anomaly description is observed in \cref{fig:xd2} and \cref{fig:xd3}.\\

\begin{figure*}[h]
  \centering
   \includegraphics[width=0.95\linewidth]{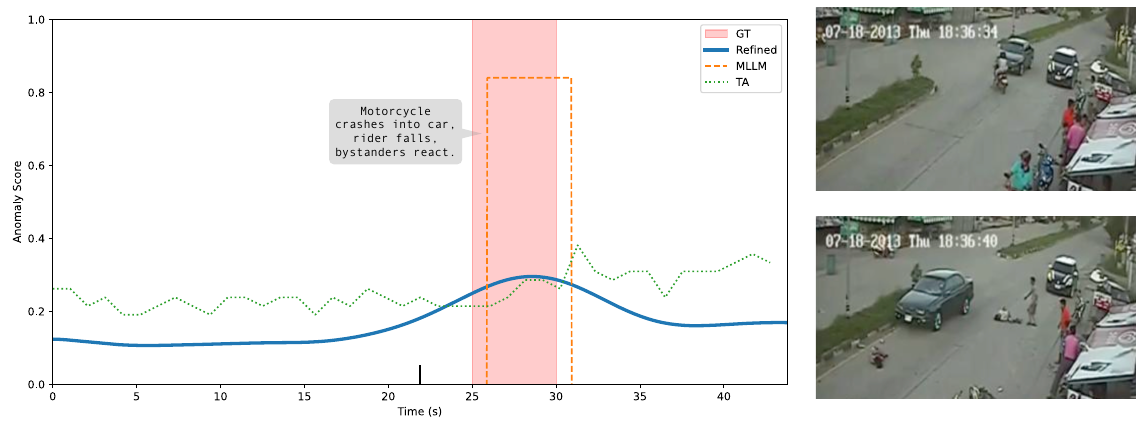}
   \caption{Results on \textit{RoadAccidents019\_x264} from UCF-Crime.}
   \label{fig:RoadAccidents019}
\end{figure*}

\begin{figure*}[h]
  \centering
   \includegraphics[width=0.95\linewidth]{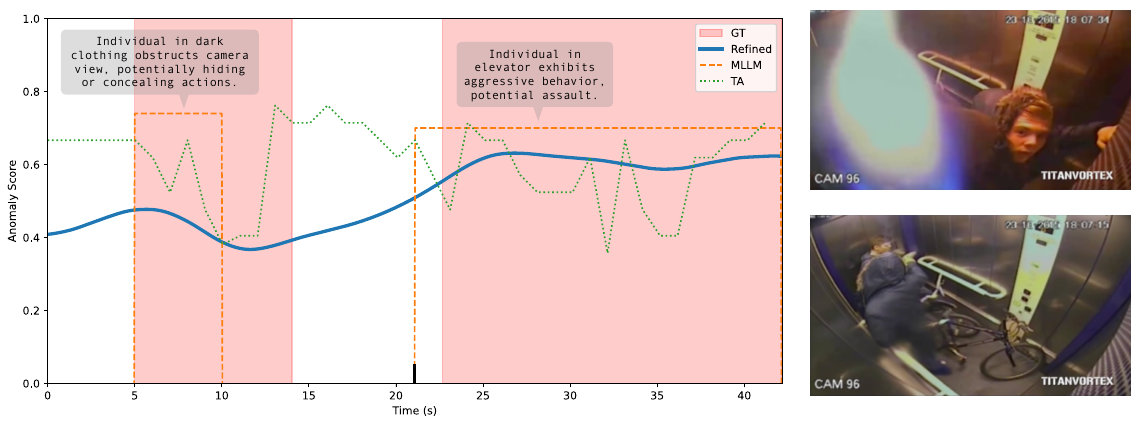}
   \caption{Results on \textit{Arson011\_x264} from UCF-Crime.}
   \label{fig:Arson011}
\end{figure*}

\Cref{fig:RoadAccidents019} and \cref{fig:Arson011} provide additional examples of how MLLM predictions refine and complement coarse anomaly curves. While textual anchors fail to capture the onset of the road accident or highlight the most critical arson moments, the MLLM predictions not only localize anomalies more precisely in time but also provide faithful descriptions of the events. For instance, in the arson video, PrismVAU flags not only the main anomalous event but also related aggressive behaviors.\\

\begin{figure*}[h]
  \centering
   \includegraphics[width=0.95\linewidth]{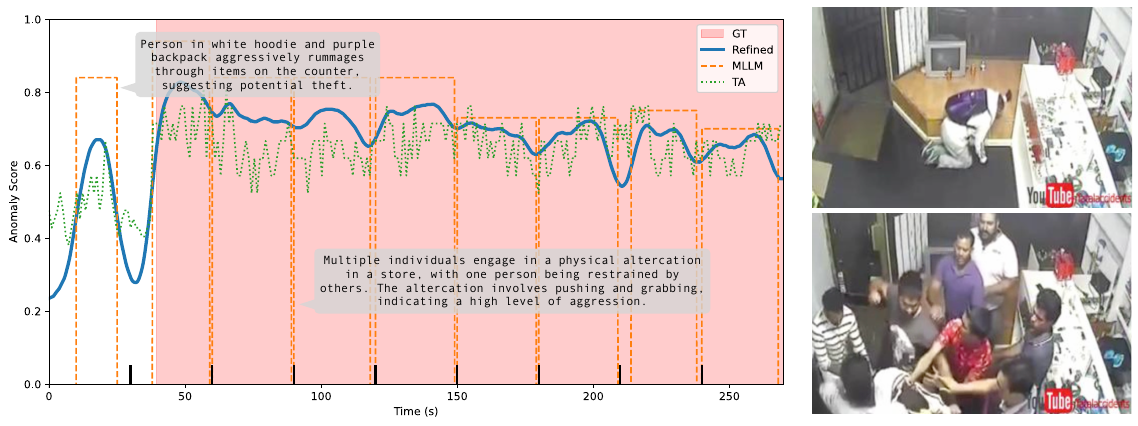}
   \caption{Results on \textit{Assault006\_x264} from UCF-Crime.}
   \label{fig:Assault006}
\end{figure*}

\begin{figure*}[h]
  \centering
   \includegraphics[width=0.97\linewidth]{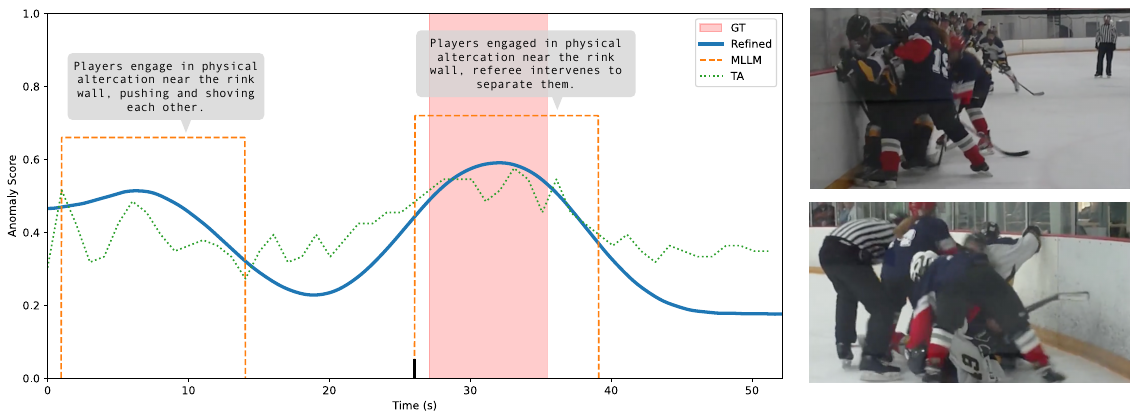}
   \caption{Results on \textit{v=pMtu7fOHdII\_\_\#1\_label\_B1-0-0} from XD-Violence.}
   \label{fig:xd4}
\end{figure*}

\begin{figure*}[h]
  \centering
   \includegraphics[width=0.95\linewidth]{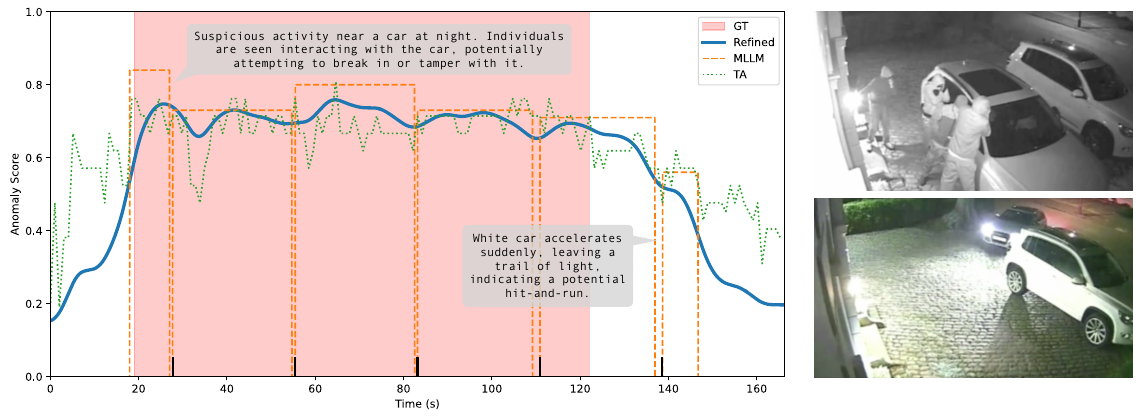}
   \caption{Results on \textit{Stealing058\_x264} from UCF-Crime.}
   \label{fig:Stealing058}
\end{figure*}

The ability to detect subtle, non-annotated criminal behaviors is also showcased in~\cref{fig:Assault006} and \cref{fig:xd4}. In the former, only the assault is annotated in the ground truth. Nonetheless, both modules of PrismVAU identify the triggering theft that precedes the assault. In the latter, an initial aggressive interaction between hockey players is flagged several seconds before the final altercation. \Cref{fig:Stealing058} further illustrates this capability: not only is the car theft detected, but also the subsequent escape, including reckless driving.\\

\begin{figure*}[h]
  \centering
   \includegraphics[width=0.95\linewidth]{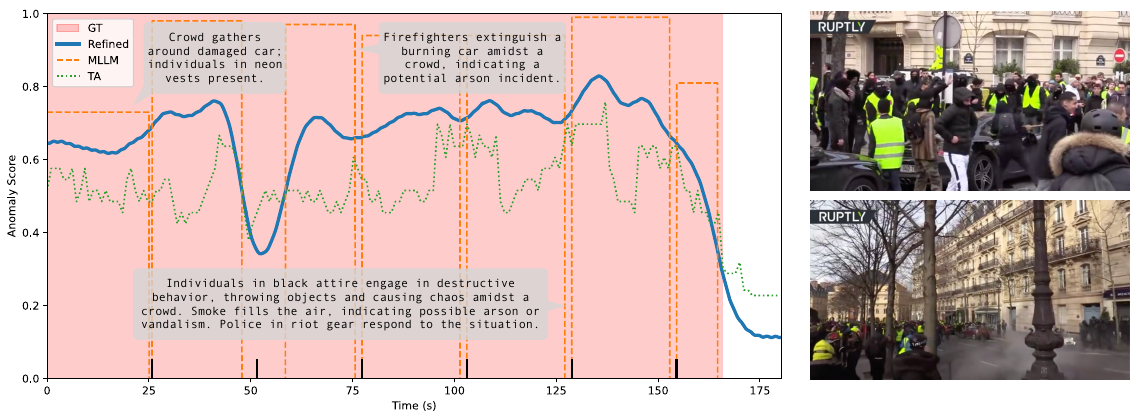}
   \caption{Results on \textit{v=ICnreR1hxP0\_\_\#1\_label\_B4-0-0} from XD-Violence.}
   \label{fig:xd1}
\end{figure*}

\begin{figure*}[h]
  \centering
   \includegraphics[width=0.97\linewidth]{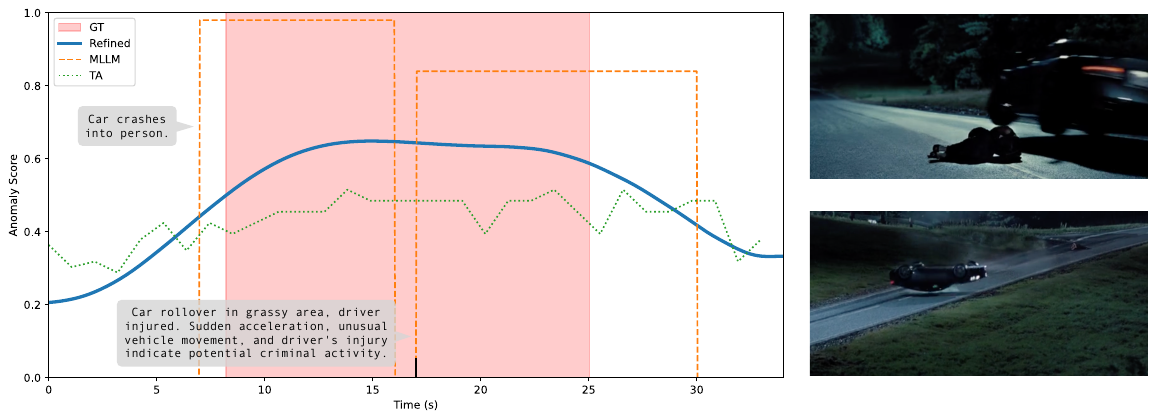}
   \caption{Results on \textit{Casino.Royale.2006\_\_\#01-46-40\_01-47-14\_label\_B6-0-0} from XD-Violence.}
   \label{fig:xd5}
\end{figure*}

\Cref{fig:xd1} highlights PrismVAU’s ability to capture a variety of distinct anomalous activities within a single riot scenario, including car damage, arson, vandalism, and police confrontation. Finally, in ~\cref{fig:xd5}, although the MLLM provides a slightly inaccurate description/cause (the car crashes while avoiding a person but does not actually hit it), it correctly identifies the consequences of the accident. \\

Additionally, \cref{fig:mllm_interaction_2} and \cref{fig:mllm_interaction_3} depict interactive conversations after PrismVAU's initial response. This interactive chat interface capability lets users engage with the model to audit system errors, elaborate on correct predictions, and explore counterfactual or hypothetical scenarios. 

\begin{figure*}
  \centering
   \begin{tcolorbox}[colback=white, colframe=orange, boxrule=1pt, width=\linewidth]
        \begingroup
        \scriptsize\ttfamily
        \textbf{PrismVAU:} \{"abnormality": 0.00, "description": "No visible criminal behavior detected in the video footage."\}\\
        
        \textbf{User:} Describe the scene and explain why it is considered normal.\\
        
        \textbf{PrismVAU:} The video shows two individuals entering an elevator in a building lobby. The scene is considered normal because there are no unusual or anomalous behaviors, such as sudden acceleration, unusual crowd clustering, suspicious objects, or contextual mismatches that could indicate criminal activity. The individuals enter the elevator and the doors close. The scene concludes with the elevator moving upwards.\\
        
        \textbf{User:} What if the individuals were running towards the elevator instead of walking?\\
        
        \textbf{PrismVAU:}  If the individuals were running towards the elevator instead of walking, it could be considered an abnormality. The sudden acceleration and unusual motion patterns would indicate a potential emergency or criminal activity, such as an attempt to escape or a robbery. In this case, the abnormality score would be higher, and further analysis would be needed to determine the specific nature of the anomaly.\\
        
        \endgroup
    \end{tcolorbox}

   \caption{Interactive dialogue with PrismVAU on a normal scene, \textit{Normal\_Videos\_783\_x264} from UCF-Crime.}
   \label{fig:mllm_interaction_2}
\end{figure*}

\begin{figure*}
  \centering
   \begin{tcolorbox}[colback=white, colframe=orange, boxrule=1pt, width=\linewidth]
        \begingroup
        \scriptsize\ttfamily
        \textbf{PrismVAU:} \{"start\_time": 10.00, "end\_time": 19.00, "abnormality": 0.85, "description": "White van collides with motorcyclist, causing fire. Motorcyclist falls, and another rider rushes to assist."\}\\
        
        \textbf{User:} Which vehicle’s actions led to the accident?\\
        
        \textbf{PrismVAU:} The white van's actions led to the accident.\\
        
        \textbf{User:} What could the drivers have done differently to avoid the collision?\\
        
        \textbf{PrismVAU:} The drivers could have maintained a safe following distance, stayed alert to their surroundings, and been prepared to react to unexpected situations on the road. Regular vehicle maintenance, such as checking brakes and tires, can also prevent accidents.\\

        \textbf{User:} What would have happened if the white van continued straight instead of turning?\\

        \textbf{PrismVAU:}  If the white van continued straight instead of turning, it is likely that the accident with the motorcyclist would have been avoided.\\
        
        \endgroup
    \end{tcolorbox}

   \caption{Interactive dialogue with PrismVAU on an abnormal scene, \textit{RoadAccidents133\_x264} from UCF-Crime.}
   \label{fig:mllm_interaction_3}
\end{figure*}